\pdfoutput=1
\documentclass[11pt]{article}

\usepackage[preprint]{acl}

\usepackage{times}
\usepackage{latexsym}

\usepackage[T1]{fontenc}
\usepackage[utf8]{inputenc}

\usepackage{microtype}

\usepackage{inconsolata}

\usepackage{graphicx}

\usepackage{enumitem}
\usepackage{amsmath,amssymb,amsfonts}
\setlist[itemize]{itemsep=2pt, parsep=2pt, topsep=2pt}
\usepackage{tcolorbox}
\tcbuselibrary{breakable}
\usepackage{CJKutf8}

\title{Internal Chain-of-Thought: Empirical Evidence for Layer‑wise Subtask Scheduling in LLMs}

\author{
 \textbf{Zhipeng Yang\textsuperscript{1, 2, 3}},
 \textbf{Junzhuo Li\textsuperscript{1, 2}},
 \textbf{Siyu Xia\textsuperscript{3}},
 \textbf{Xuming Hu\textsuperscript{1, 2, $\dagger$}}
\\
\\
 \textsuperscript{1}The Hong Kong University of Science and Technology (Guangzhou),\\
 \textsuperscript{2}The Hong Kong University of Science and Technology,\\
 \textsuperscript{3}Southeast University
}

\begin{document}
\maketitle

\begin{abstract}
We show that large language models (LLMs) exhibit an \textit{internal chain-of-thought}: they sequentially decompose and execute composite tasks layer-by-layer. Two claims ground our study: (i) distinct subtasks are learned at different network depths, and (ii) these subtasks are executed sequentially across layers. On a benchmark of 15 two-step composite tasks, we employ layer-from context-masking and propose a novel cross-task patching method, confirming (i). To examine claim (ii), we apply LogitLens to decode hidden states, revealing a consistent layerwise execution pattern. We further replicate our analysis on the real-world \textsc{Trace} benchmark, observing the same stepwise dynamics. Together, our results enhance LLMs transparency by showing their capacity to internally plan and execute subtasks (or instructions), opening avenues for fine-grained, instruction-level activation steering.
\end{abstract}

\begingroup
\renewcommand\thefootnote{}\footnotetext{$^\dagger$~Corresponding author.}
\addtocounter{footnote}{-1}
\endgroup

\begingroup
\renewcommand\thefootnote{}\footnotetext{Code available at: \url{https://github.com/yzp11/Internal-Chain-of-Thought}}
\addtocounter{footnote}{-1}
\endgroup

\section{Introduction}
Large Language Models (LLMs) excel at solving complex tasks such as instruction following, and multi-step problem solving \cite{zhang2024iopo, zengevaluating, wang2024llm}. Much recent progress relies on explicit ``chain of thought'' \cite{wei2022chain, zhangautomatic}, which guides models to decompose multi-step problems into intermediate reasoning stages. This raises a foundational question: \textbf{Do LLMs also perform such multi-step reasoning internally, without revealing steps in their output?} In this work, we answer \textbf{yes}: LLMs exhibit an internal chain-of-thought (ICoT), meaning they internally break down composite tasks and process their components sequentially across network layers. Going beyond interpretability studies on latent factual multi-hop reasoning \cite{yang2024large, biran2024hopping, yu2025back, biology}, we investigate task-level reasoning rather than just chains of facts.

\begin{figure}[t]
  \centering
  \includegraphics[width=\columnwidth]{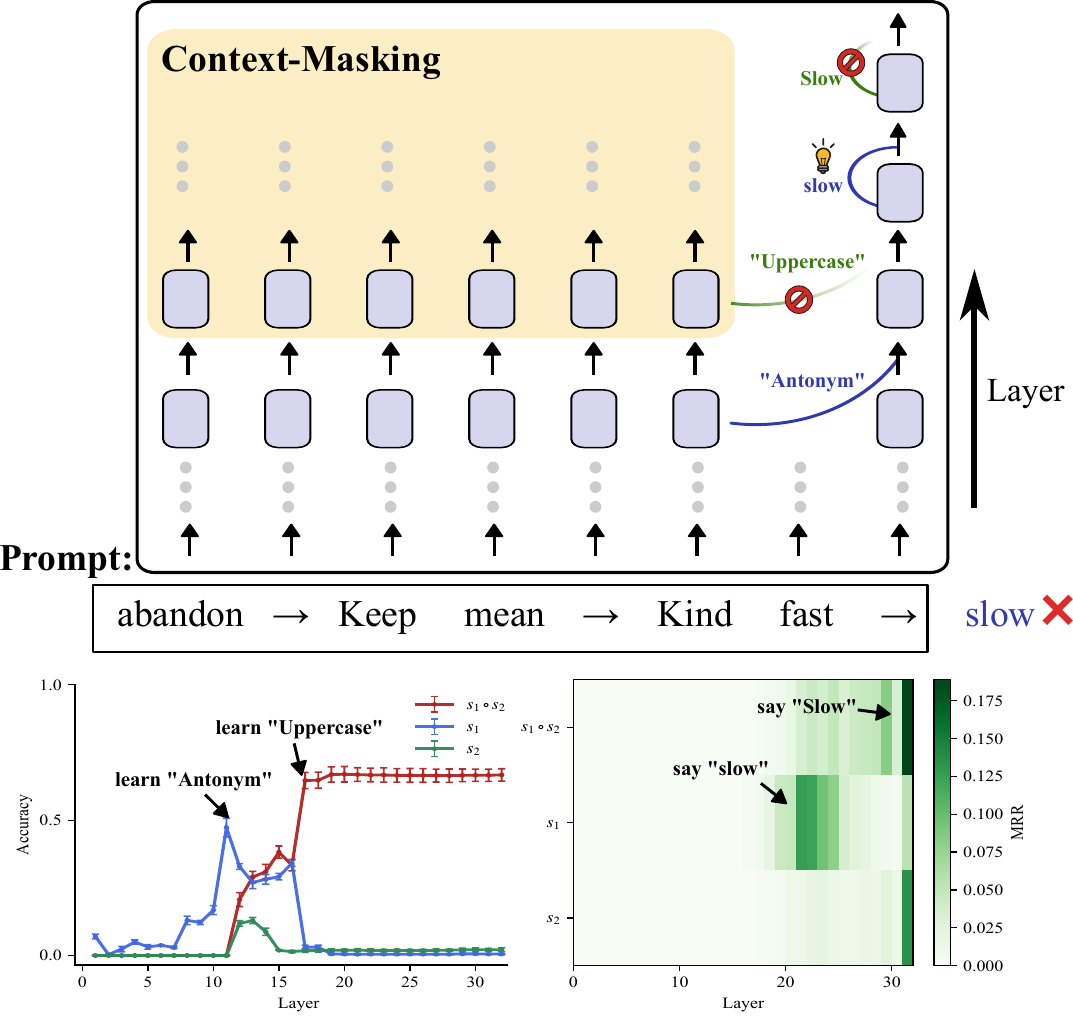}
  \caption{\label{diagram}
  Evidence of an internal chain-of-thought. Selectively masking (bottom left) from specific layers can preserve the first subtask (antonym) while ablating the second (uppercase). Decoding hidden states (bottom right) on a clean run shows the intermediate answer (``\texttt{slow}'') peaks at middle layers.}
\end{figure}

To illustrate this concept, consider a composite task that requires two steps: antonym then uppercase. For example, given the input ``\texttt{fast}'', solving this involves an intermediate step—finding the antonym ``\texttt{slow}''—followed by capitalizing it to ``\texttt{Slow}'' (with S capitalized). If an internal chain-of-thought exists, we would expect the hidden state at some intermediate layer to represent ``\texttt{slow}'', and later layers to transform it into ``\texttt{Slow}''. In general, the presence of an ICoT implies that distinct phases of computation occur inside the model, each corresponding to a subtask in the overall problem. Figure \ref{diagram} illustrates our core findings: selectively masking context from specific layers can preserve the first subtask (antonym) while ablating the second (uppercase). Meanwhile, decoding results of hidden state on clean run show that intermediate answer (``\texttt{slow}'') often peaks at middle layers.

For tractability, we study tasks that decompose into two sequential subtasks (denoted $t := s_1 \circ s_2$). The task $t$ maps an input $x$ to an output $y^t = s_2(s_1(x))$. We frame our analysis through Task Vector framework \cite{hendel2023context, toddfunction, liimplicit, saglam2025learning}, which divides in-context learning (ICL) into two phases: (1) a \textit{learning phase}, where the model abstracts task rules into a hidden representation, task vector, and (2) a \textit{rule application phase}, where the query is processed using this vector. For a model $T$, given demonstrations $S$ and a query $x$, the process is modeled as:
\begin{equation}
\label{ICL}
    T\left( \left[S, x\right]\right) \Rightarrow \underbrace{\theta = \mathcal{A}\left(S\right)}_{\mathrm{learning\ phase}}, \underbrace{y=f\left(x; \theta\right)}_{\mathrm{application\ phase}},
\end{equation}
where $\mathcal{A}$ abstracts the task vector $\theta \in \mathbb{R}^d$, and $f$ denotes the application of task-specific rules. Extending to composite tasks yields two claims:
\begin{itemize}
\item \textbf{Claim 1.} Subtasks are learned at different network depths, inducing an intermediate subtask vector (either $\theta^{s_1}$ or $\theta^{s_2}$) that generalizes to its corresponding subtask.
\item \textbf{Claim 2.} Subtasks are executed sequentially across layers. At depth $l_1$, the model applies $f^{l_1}(x; \theta^{s_1})$ to compute the first subtask. At a later depth $l_2$, it applies $f^{l_2}(y^{s_1}; \theta^{s_2})$, yielding the final result.
\end{itemize}
Crucially, we distinguish between two processes: ``learning'' (Claim 1) and ``execution'' (Claim 2), corresponding to the learning phase and application phase in Task Vector framework, respectively.

We first introduce a benchmark of 15 two-step composite tasks spanning four categories (Section \ref{section::dataset}). We present two lines of evidence for Claim 1: (1) layer-from context-masking \cite{sia2024does}, which blocks attention to demonstrations after layer $l$ to reveal where each subtask is learned, and (2) cross-task patching, a novel method which inserts residual activations from a composite prompt into zero-shot sub-task queries to detect reusable ``subtask vectors''. Across four models and 15 two-step tasks, masking reveals a sharp ``X-shape'' (Figure \ref{mask_main}), indicating a sequential learning dynamics:  the model first abstracts the rule for $s_1$ at an earlier layer, and later learns $s_2$ at a deeper layer. Meanwhile, patching activations in Llama-3.1-8B (see Table \ref{subtask-main}) yield transferable subtask vectors to a significant degree (66\% on average).

Next, to verify Claim 2, We decode every layer with LogitLens \cite{logitlen}, projecting hidden states into token space and tracking the mean reciprocal rank of the first-step target ($y^{s_1}$ or $y^{s_2}$) versus the final answer ($y^{s_1 \circ s_2}$). Decoding results show the same ``handoff'' (see Figure \ref{decoding_components} and \ref{decoding_res}): intermediate answer peaks in mid-layers, then is overtaken a few layers later by the final answer. Finally, we replicate layer-from context-masking on \textsc{TRACE} \cite{zhang2024iopo}, a complex instruction-following benchmark, demonstrating that the same sequential learning dynamics emerge in real-world settings (see Figure \ref{trace}). The primary contributions of this study are as follows:
\begin{itemize}
\item We construct a curated benchmark of 15 composite tasks spanning four categories.
\item We employ context-masking and propose \textbf{cross-task patching}, demonstrating that subtasks are learned at different depths, inducing an intermediate subtask vector.
\item We use LogitLens to decode hidden states, revealing a consistent layerwise execution pattern.
\item We replicate our method on the \textsc{TRACE} benchmark, confirming the same finding also emerge in practical settings.
\end{itemize}

Our findings enhance LLM transparency by revealing their capacity to internally plan and execute subtasks (or instructions). This aligns with, and extends, prior interpretability studies on multi-hop reasoning \cite{yang2024large, biran2024hopping, yu2025back, biology} and look-ahead planning \cite{men2024unlocking}. While those often focus on factual recall or predictive steps, our work investigates task-level reasoning rather than just chains of facts. Furthermore, the discovery of ICoT opens exciting avenues for fine-grained, instruction-level behavior control. For instance, by identifying the layers responsible for processing specific (potentially harmful) instructions within a user's prompt, we could directionally intervene to steer their execution for safer LLM behavior.

\begin{table*}
  \centering
  \begin{tabular}{lll}
    \hline
    \textbf{Category} & \textbf{Task Description} & \textbf{Example (Input $\rightarrow$ Output)} \\
    \hline
    Knowledge–Algorithmic &
    \begin{tabular}[c]{@{}l@{}}$s_1$: antonym \\ $s_2$: uppercase\end{tabular} &
    \texttt{fast} $\rightarrow$ \texttt{Slow} \\
    \hline
    Extractive–Knowledge &
    \begin{tabular}[c]{@{}l@{}}$s_1$: select adjective \\ $s_2$: synonym\end{tabular} &
    \texttt{artistic, captain, bring} $\rightarrow$ \texttt{creative} \\
    \hline
    Extractive–Algorithmic &
    \begin{tabular}[c]{@{}l@{}}$s_1$: select last item \\ $s_2$: first letter\end{tabular} &
    \texttt{spicy, cowardly, hoop} $\rightarrow$ \texttt{h} \\
    \hline
    Knowledge–Translation &
    \begin{tabular}[c]{@{}l@{}}$s_1$: retrieve country \\ $s_2$: translate to French\end{tabular} &
    \texttt{Cenepa River} $\rightarrow$ \texttt{Pérou} \\
    \hline
  \end{tabular}
  \caption{\label{task_example}
    Representative examples from the composite task benchmark across four categories. Each task involves a sequential application of two subtasks, though no intermediate outputs are shown in-context. See Appendix \ref{appendix-dataset} for a complete task list and definitions.
  }
\end{table*}

\section{Experimental Setup}
\subsection{Prompt Design}
We focus on composite tasks that naturally decompose into two sequential subtasks—for example, retrieving domain knowledge and then translating it, or extracting information followed by format transformation. Formally, we represent a composite task as $t := s_1 \circ s_2$, where $s_1$ and $s_2$ are sequentially applied subtasks. Given a query $x$, the final output is $y^t = s_2(s_1(x))$. For analysis purposes, we also compute intermediate outputs corresponding to the isolated application of each subtask: $y^{s_1} = s_1(x)$ and $y^{s_2} = s_2(x)$. As an illustrative example, consider $s_1 = \text{``antonym''}$ and $s_2 = \text{``uppercase''}$. For input $x = \texttt{``fast''}$, the correct intermediate and final outputs would be $y^{s_1} = \texttt{``slow''}$, $y^{s_2} = \texttt{``Fast''}$, and $y^t = \texttt{``Slow''}$.

For each composite task $t \in \mathcal{T}$ in our task suite $\mathcal{T}$, we construct a dataset $\mathcal{P}_t$ consisting of in-context prompts $p_i^t \in \mathcal{P}_t$. Each prompt includes $N$ input-output demonstration pairs of the form $(x, y^t)$, showing the full composite transformation, followed by a query input $x_{iq}$ for which the model is expected to predict the corresponding target $y_{iq}^t$. Notably, no intermediate outputs or reasoning steps are included in the prompt. The in-context learning (ICL) prompt format is (Prompt details can be found in Appendix \ref{appendix-prompt}):
\begin{equation}
    p_i^t = \left[(x_{i1}, y_{i1}^{s_1 \circ s_2}), \dots, (x_{iN}, y_{iN}^{s_1 \circ s_2}), x_{iq} \right].
\end{equation}

\begin{figure*}[htbp]
	\centering
	\includegraphics[width=16cm]{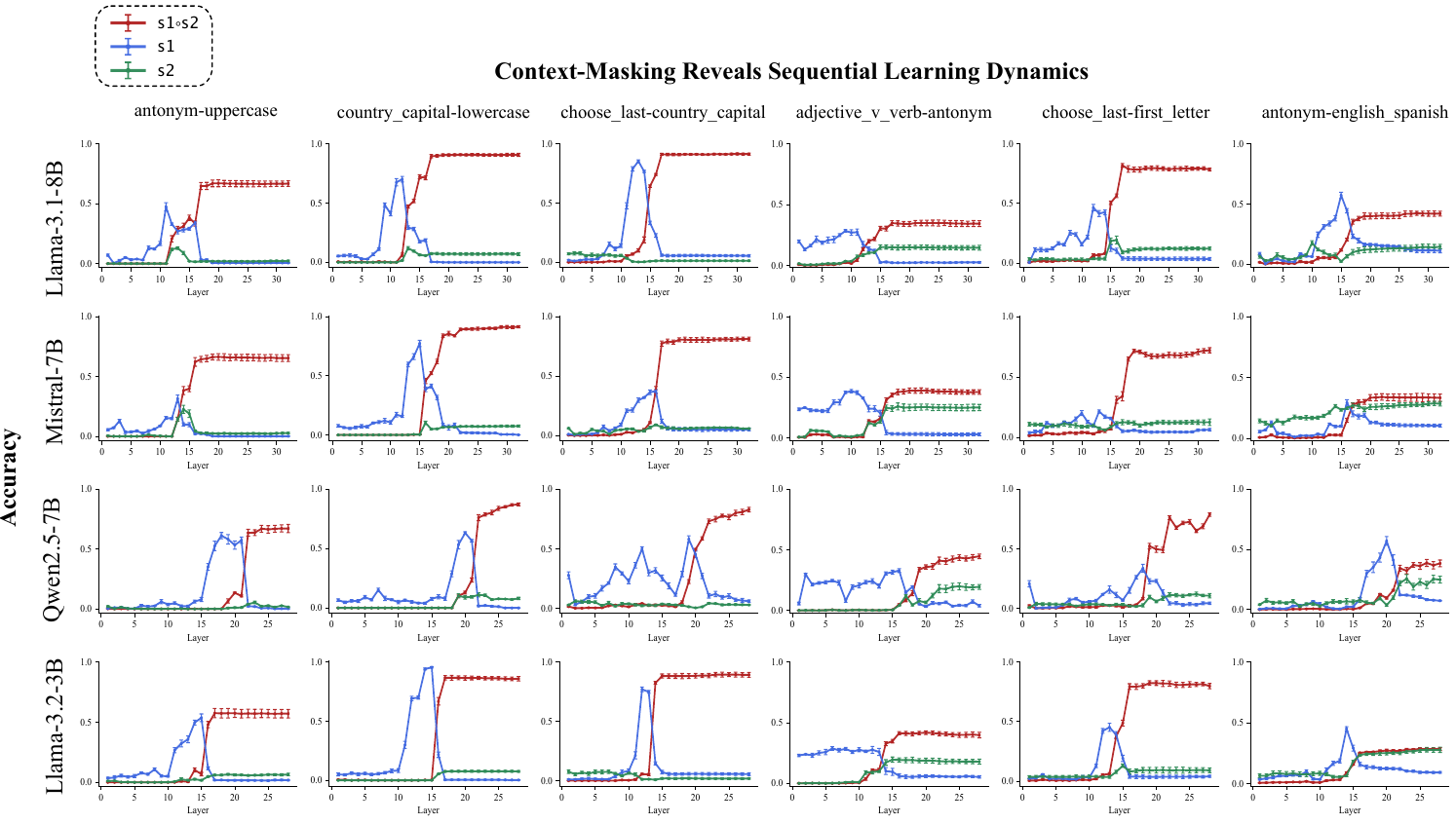}
	\caption{\label{mask_main}  Layer-from context-masking results for six composite tasks across four models. The ``X-shape'' pattern reveals sequential learning dynamics.}
\end{figure*}

\subsection{Dataset}
\label{section::dataset}
We construct a benchmark of 15 composite tasks spanning four categories:
\begin{itemize}
    \item \textbf{Knowledge–Algorithmic:} Tasks that combine factual knowledge retrieval (e.g., country capitals, antonyms) with deterministic transformations (e.g., uppercase conversion).
    \item \textbf{Extractive–Knowledge:} Tasks that require identifying items from a list (e.g., selecting the last item) followed by a knowledge-based operation (e.g., finding a related concept).
    \item \textbf{Extractive–Algorithmic:} Tasks that involve list-based selection followed by symbolic transformations (e.g., case conversion, character extraction).
    \item \textbf{Knowledge–Translation:} Tasks that combine knowledge retrieval with language translation (e.g., translating the capital city of a given country into French or Spanish).
\end{itemize}
Each query in the dataset requires the sequential execution of two subtasks in a fixed order as defined by the task specification. However, we do not assume that LLMs necessarily follow this order during internal processing. To probe the latent execution path, we measure intermediate outputs corresponding to the isolated application of each subtask: $y^{s_1} = s_1(x)$ and $y^{s_2} = s_2(x)$. Table \ref{task_example} presents illustrative examples for each category. Full details and descriptions for all 15 composite tasks can be found in Appendix \ref{appendix-dataset}.

\section{Background}
We consider an autoregressive transformer language model $T$ that takes an input prompt $p$ and outputs a next-token distribution $T(p)$ over a vocabulary $\mathcal{V}$. Internally, $T$ consists of $L$ transformer layers connected via a residual stream \cite{elhage2021mathematical}. We focus our analysis on the residual stream at the final token position. Embedding matrix $\mathbf{W}_E \in \mathbb{R}^{|\mathcal{V}|\times d}$ first maps the last token to a hidden representation as initial residual stream $\mathbf{h}^0 \in \mathbb{R}^d$. At each layer $l$, the model adds the outputs of the self-attention and feedforward network (FFN) modules to the residual stream from the previous layer.  Formally, the residual stream at layer $l$ is given by:
\begin{equation}
    \mathbf{h}^l = \mathbf{h}^{l-1} + \mathbf{A}^l + \mathbf{F}^l,
\end{equation}
where $\mathbf{A}^l \in \mathbb{R}^d$ and $\mathbf{F}^l \in \mathbb{R}^d$ denote the attention and FFN outputs at the final token position, respectively, at layer $l$. We adopt a simplified formulation here for clarity; in practice, additional components such as Layer Normalization are also applied.

\section{Claim 1: Intermediate Subtask Representations}
We present two lines of evidence for \textbf{Claim 1}: (1) using layer-from context-masking, we show that different layers are responsible for learning each subtask; (2) with a novel cross-task patching method, we demonstrate that subtask-specific vectors emerge at the final token position, serving as abstract representations that generalize across tasks.

\subsection{Layer-from Context-Masking}
\label{section-masking}
In-Context Learning requires model to infer a task from examples and apply it to a new input, as formalized in Equation \ref{ICL}. If LLMs follow a internal chain-of-thought, then each subtask in a composite task should be learned at a distinct point in the network. That is, subtask learning should unfold sequentially across layers—rather than all at once—making intermediate learning states observable. To investigate this, we employ layer-from context-masking \cite{sia2024does}. This technique disables access to the in-context examples (task demonstrations) from a specific layer onward by masking all attention to context tokens. If masking is applied from the input layer ($l=0$), the model cannot attend to any demonstrations, and ICL should fail. However, if masking begins only after the model has learned the task, then its performance should remain intact. Crucially, by gradually shifting the start-masking layer from early to late, we can infer the sequential dynamics of the model’s learning process.

Let $\mathbf{A} = \frac{\mathbf{Q}\mathbf{K}^\top}{\sqrt{D}}$ denote the raw attention scores in a decoder-only transformer, where $\mathbf{Q}$ and $\mathbf{K}$ are the query and key matrices, respectively, and $D$ is the dimensionality of the hidden states. For token positions $i$ and $j$, the element $A_{ij}$ represents how much token $i$ attends to token $j$. We apply a context masking to disable attention to in-context examples, as $A_{ij}+m(j,U)$. The mask $m(j, U)$ is defined as:
\begin{equation}
    m(j, U) =
    \begin{cases}
        0 & \text{if } j \notin U, \\
        -\infty & \text{if } j \in U,
    \end{cases}
\end{equation}
where $U$ is the set of indices of all in-context example tokens. The mask is applied from layer $l$ onward, such that for all $l' \geq l$, attention to context is zeroed out after Softmax. For each test prompt, we progressively increase the masking layer $l$ from $1$ to $L$ and record the model’s prediction accuracy on both intermediate and final outputs. For composite tasks, we aim to identify a two-phase masking pattern. At early layers, masking may lead the model to predict intermediate outputs—e.g., $y^{s_1}$ or $y^{s_2}$—indicating that the model has learned the first subtask but not yet the second. As masking is delayed to deeper layers, the model's predictions should transition sharply from intermediate answers to the final answer $y^t$, revealing a layered acquisition of subtasks. By contrast, if the model transitions directly from generating no meaningful output to the correct final answer as masking depth increases, without producing intermediate completions, this would suggest a monolithic in-context learning process. This distinction is central to testing whether subtask learning unfolds sequentially across layers.

\begin{table}[t]
  \centering
  \renewcommand{\arraystretch}{1.2}
  \setlength{\tabcolsep}{6pt}
  \begin{tabular}{p{4.1cm}c}
    \hline
    \textbf{Composite Task} & \textbf{Llama-3.1-8B} \\
    \hline
    \texttt{antonym-uppercase} &  \\
    \quad $s_1$ (antonym) & 0.92 $\pm$ 0.02 \\
    \quad $s_2$ (uppercase) & 0.24 $\pm$ 0.03 \\
    
    \texttt{country\_capital-lowercase} & \\
    \quad $s_1$ (country\_capital) & 0.88 $\pm$ 0.03 \\
    \quad $s_2$ (lowercase) & 0.49 $\pm$ 0.05 \\
    
    \texttt{choose\_last-country\_capital} & \\
    \quad $s_1$ (choose\_last) & 0.44 $\pm$ 0.03 \\
    \quad $s_2$ (country\_capital) & 0.98 $\pm$ 0.02 \\
    
    \texttt{adjective\_v\_verb-antonym} & \\
    \quad $s_1$ (adj\_v\_verb) & 0.38 $\pm$ 0.11 \\
    \quad $s_2$ (antonym) & 0.94 $\pm$ 0.03 \\
    
    \texttt{choose\_last-first\_letter} & \\
    \quad $s_1$ (choose\_last) & 0.29 $\pm$ 0.03 \\
    \quad $s_2$ (first\_letter) & 1.01 $\pm$ 0.01 \\
    
    \texttt{antonym-english\_spanish} & \\
    \quad $s_1$ (antonym) & 0.91 $\pm$ 0.02 \\
    \quad $s_2$ (english\_spanish) & 0.45 $\pm$ 0.03 \\

    \hline
    \textbf{Average} & 0.66\\
    \hline
  \end{tabular}
  \caption{\label{subtask-main}
    Subtask vector strength for six representative composite tasks in Llama-3.1-8B. Each composite task’s average residual activation is patched into each subtask, with results shown for $s_1$ and $s_2$ respectively.}
\end{table}

\paragraph{Experiment.}
We conduct our layer-from context-masking analysis on four LLMs: Llama-3.1-8B \cite{grattafiori2024llama}, Mistral-7B \cite{jiang2024mistral}, Qwen2.5-7B \cite{yang2024qwen2}, and Llama-3.2-3B \cite{llama32}, evaluating their behavior across all 15 composite tasks. We selected models in the 3B–8B parameter range to ensure broad coverage across popular, open-source checkpoints while maintaining interpretability and accessibility for analysis. Larger models were excluded due to resource constraints and the increased difficulty of probing internal representations. For each task, we generate 500 test prompts, sampled uniformly at random from the corresponding dataset. Each prompt includes $N$ in-context examples (following prior work \cite{hendel2023context}, we set $N = 5$). To ensure robustness, all experiments are repeated across five random seeds, and we report averaged results. We mask all tokens from the in-context examples—including both content tokens and ``template'' tokens such as separators (Q:, A:, newlines, etc.).

In the resulting sequential learning dynamics plots (see Figure \ref{mask_main} and Appendix \ref{appendix-masking}), we observe a striking ``X-shape'' pattern across most composite tasks, with a few exceptions (e.g., \texttt{choose\_last–landmark\_country} in Llama-3.1-8B). Specifically, as context masking is delayed to deeper layers, the model's output transitions from generating one of the intermediate answers (e.g., the result of $s_1$) to producing the correct final answer. The intersection point—where performance on the intermediate answer begins to drop while performance on the final answer rises—suggests a boundary between subtask learning phases. This structure provides compelling evidence for sequential learning dynamics: the model first abstracts the rule for $s_1$ at an earlier layer, and later learns $s_2$ at a deeper layer. 

\subsection{Cross-Task Patching}
While context-masking reveals when subtask information is acquired, it does not directly test whether LLMs represent individual subtasks as reusable, abstract vectors. To address this, we introduce \textbf{cross-task patching}, a novel method that investigates whether sequential learning dynamics produce intermediate \textit{subtask vectors}. Prior work suggests that the residual stream at the final token position encodes a latent task representation $\theta$ derived from in-context examples \cite{hendel2023context, liimplicit, toddfunction}. These representations can be replaced into the hidden states while running model on other prompts to influence model behavior. Here, we extend this idea to composite tasks. Specifically, we examine whether the activations obtained from a composite prompt can be used to improve performance on each subtask individually. We compute the average residual stream activation across composite task prompts, then patch it into zero-shot prompts from the subtask datasets. If performance on the subtask improves, we infer that the composite prompt's activation encodes the corresponding subtask vector.

Formally, we begin by running the model on a set of composite prompts $p^t_i \in \mathcal{P}_t$, each containing $N$ examples of task $t$, and extract activation vector at the final token position from each layer $l$. Averaging over all prompts yields a layerwise task representation:
\begin{equation}
    \bar{\mathbf{h}}^t_l = \frac{1}{|\mathcal{P}_t|} \sum_{p_i^t \in \mathcal{P}_t} \mathbf{h}^l(p_i^t).
\end{equation}
We then patch this vector into a set of zero-shot subtask prompts $\tilde{p}_i \in \tilde{\mathcal{P}}_{s_j}$ (i.e., prompts with no in-context examples), replacing the residual stream at layer $l$ with $\bar{\mathbf{h}}^t_l$, and evaluate the model's performance:
\begin{equation}
    \mathrm{Acc}(\tilde{\mathcal{P}}_{s_j}, l) = \frac{1}{|\tilde{\mathcal{P}}_{s_j}|} \sum_{\tilde{p}_i} \mathbb{I}\left[T(\tilde{p}_i \mid \mathbf{h}^l := \bar{\mathbf{h}}^t_l) = y_i\right].
\end{equation}
To quantify how well this patched vector recovers the subtask behavior, we define a normalized \textit{subtask vector strength}. As the patching is used on each layer, we choose the best result to calculate the subtask vector strength:
\begin{equation}
    \mathrm{Strength}^{s_j} = \frac{\max_l \mathrm{Acc}(\tilde{\mathcal{P}}_{s_j}, l) - \mathrm{Acc}(\tilde{\mathcal{P}}_{s_j})}{\mathrm{Acc}(\mathcal{P}_{s_j}) - \mathrm{Acc}(\tilde{\mathcal{P}}_{s_j})},
\end{equation}
where $\mathrm{Acc}(\mathcal{P}_{s_j})$ is the subtask's performance under standard ICL (with $N$ examples), and $\mathrm{Acc}(\tilde{\mathcal{P}}_{s_j})$ is the zero-shot baseline. A strength of 1 implies full recovery of subtask performance, indicating a fully formed subtask vector; a strength of 0 implies no transfer. We test the subtask vector strength on both subtasks $s_1$ and $s_2$.

While our method is inspired by \cite{hendel2023context}, it differs in both objective and application. \citet{hendel2023context} copy activations between prompts within the same task to test whether task information is encoded in the residual stream. In contrast, our cross-task patching copies activations from a composite-task prompt into a subtask prompt, allowing us to ask whether subtask representations are embedded and transferable from composite-task inference.

\paragraph{Experiment.} To ensure independence between datasets, we first split each subtask dataset into disjoint train and test subsets (see Appendix \ref{appendix-dataset} for details about subtask dataset). Composite datasets are constructed using only the train set, while zero-shot patching is evaluated on held-out subtask examples. We compute $\bar{h}_l^t$ using 100 composite prompts and test patching strength on 500 zero-shot subtask prompts. We repeat this process across 15 composite tasks, 4 models, and 5 random seeds.
 
Table \ref{subtask-main} and Appendix \ref{appendix-patching} report the patching strength across tasks. We find that most composite tasks yield transferable subtask vectors to a significant degree (0.66 on average). Interestingly, all composite tasks exhibit asymmetric transfer—for instance, the composite vector may strongly support $s_1$ but only weakly support $s_2$. This asymmetry may reflect either the task type of $s_2$ (e.g., extractive tasks), or that $s_2$ is applied in a more entangled fashion atop the result of $s_1$, making its representation more context-dependent.

\section{Claim 2: Layer-wise Rule Application}
\textbf{Claim 2} hypothesizes that LLMs apply rules for composite tasks in a staged process: at an earlier layer $l_1$, the model applies a function $f^{l_1}(x; \theta^{s_1})$ to perform the first subtask; later, at layer $l_2 > l_1$, it applies a second function $f^{l_2}(y^{s_1}; \theta^{s_2})$, integrating this intermediate representation with the second subtask's logic to produce the final answer. Crucially, we should be able to trace this transformation through the model's residual stream, which accumulates the outputs of each attention and MLP block.

We decode the next-token probabilities for each intermediate layer using LogitLens \cite{logitlen}. This method aims to project hidden states into the vocabulary space. Formally, let $\mathbf{h}^l$ denote the residual stream at the final token position, at layer $l$. To decode their outputs into probability distributions $\mathbf{p}$ over vocabulary tokens, we use the unembedding matrix $\mathbf{W_U} \in \mathbb{R}^{d \times |\mathcal{V}|}$, along with a normalization that rescales component activations relative to the final-layer logits:
\begin{equation}
    \mathbf{p} = \mathrm{Softmax}\left( \mathbf{W_U} \cdot \frac{\mathbf{h}^l - \bar{\mathbf{h}}^l}{\alpha^*} \right), 
\end{equation}
where $\bar{\mathbf{h}}^l$ are the mean component outputs for normalization, and $\alpha^*$ is a scaling factor derived from the final layer's residual norm. Besides, we also decode each attention and MLP block's output $\mathbf{A}^l$ and $\mathbf{F}^l$. We then measure the \textbf{Mean Reciprocal Rank (MRR)} for three specific targets: $y^t$, $y^{s_1}$ and $y^{s_2}$.

Since the LogitLens method provides only correlational evidence, it can indicate that a composite subtask is computed later than its precursor, but not that the output of the first computation is actually used by the second. To address this limitation, we conducted an additional causal intervention experiment, targeting the residual stream at the peak layer of the subtask $s_1$ identified in our decoding analysis (e.g., Figure \ref{decoding_res}). Full details can be found in Appendix \ref{appendix-causal}.

\begin{figure}[t]
  \includegraphics[width=\columnwidth]{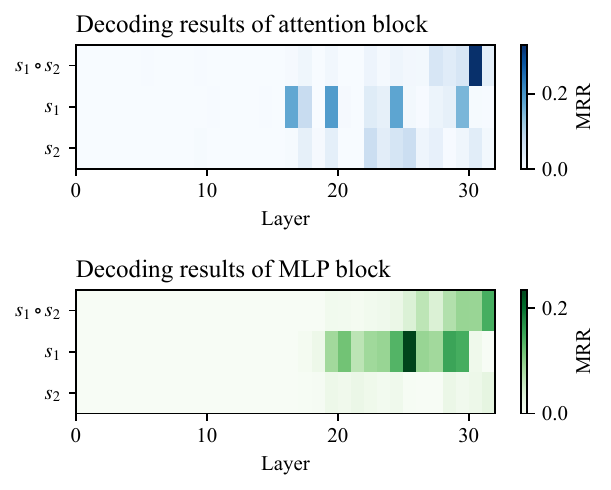}
  \caption{\label{decoding_components}Heatmaps of attention and MLP block decoding results for the \texttt{country\_capital-lowercase} task in Llama-3.1-8B.}
\end{figure}

\begin{figure}[t]
  \includegraphics[width=\columnwidth]{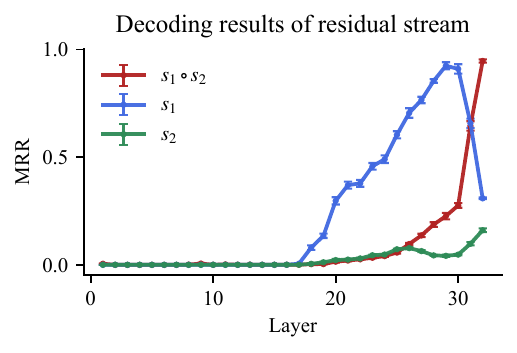}
  \caption{\label{decoding_res}Decoding results of the residual stream for the \texttt{country\_capital-lowercase} task in Llama-3.1-8B.}
\end{figure}

\paragraph{Experiment.} We conduct this analysis on 500 prompts for each of the 15 composite tasks described earlier. For each prompt, we extract and decode the attention outputs, MLP outputs, and residual stream at each layer, compute mean reciprocal ranks (MRRs) for the three target outputs described above, and plot the resulting trajectories. For MRR computation, we consider all possible vocabulary tokens, including nonsensical ones, when ranking model predictions. For multi-token answers (e.g., \textit{United Kingdom}), we evaluate only the first token (e.g., \textit{United}). This applies to both intermediate targets $s(x)$ and final outputs $s_2(s_1(x))$. We chose this approach to ensure consistency across all tasks and models.

Figure \ref{decoding_components} shows the heatmaps of attention and MLP block decoding results for the \texttt{country\_capital-lowercase} task in Llama-3.1-8B. Figure \ref{decoding_res} displays the decoding results of the residual stream (see Appendix \ref{appendix-decoding} for full results). We observe a clear layerwise task execution pattern: the model produces intermediate answers in middle layers, which are progressively surpassed by the final answer in later layers. The crossover point—where MRR for $y^{s_1}$ declines while MRR for $y^{s_1 \circ s_2}$ increases—mirrors the two-stage task execution hypothesized in Claim 2. In rarer cases such as \texttt{choose\_last–landmark\_country}, the model seems to compute the full composition at an early stage, without exhibiting a clear intermediate phase.

\section{A Practical Case: \textsc{TRACE} Dataset}
\label{section:trace}
To evaluate the applicability of our analysis in real-world scenarios, we extend our experiments to \textsc{TRACE} \cite{zhang2024iopo}, a Chinese complex instruction following benchmark. \textsc{TRACE} is built on a manually curated taxonomy of complex instructions, incorporating 26 constraint dimensions grouped into five high-level categories. Each prompt in \textsc{TRACE} consists of two components: a \textit{Task Description}, which defines the core objective (e.g., ``Introduce Adagrad''), and a set of \textit{Constraints}, which specify additional requirements that the model must satisfy. For example, a representative prompt might be (See Appendix \ref{appendix-prompt} for the original Chinese version):
\begin{tcolorbox}[breakable]
\textbf{Task Description:} Explain the Adagrad algorithm in detail, playing the role of a machine learning expert.\\
\textbf{Constraints:} 
1. Write at least 1000 words; \\
2. The explanation must include the origin, principles, pros and cons of the Adagrad algorithm, as well as its applications in practical scenarios;\\
3. Use LaTeX format to represent all mathematical formulas and algorithm steps;\\
4. Ensure that the explanation of the principles is both professional and easy to understand.
\end{tcolorbox}
We provide the entire prompt to the model and use a LLM evaluator to assign a score (from 0 to 10) for each constraint based on how well it is satisfied. In this section, we apply layer-from context-masking, but with a twist: we selectively mask only the \textit{Constraints} portion of the prompt from each layer onward, while retaining access to the \textit{Task Description} throughout. Our goal is to determine whether different types of constraints are learned at different depths, thereby exhibiting a multi-step learning trajectory. However, it does not attempt to replicate the ``X-shape'' pattern described in Section \ref{section-masking}, since \textsc{TRACE} constraints are applied in parallel rather than sequentially. Because these constraints operate simultaneously, satisfying one does not entail suppressing another. Accordingly, instead of expecting mutually exclusive transitions, we focus on differences in when and how sharply individual constraints are learned, as reflected in their scoring curves.

\begin{figure}[t]
  \includegraphics[width=\columnwidth]{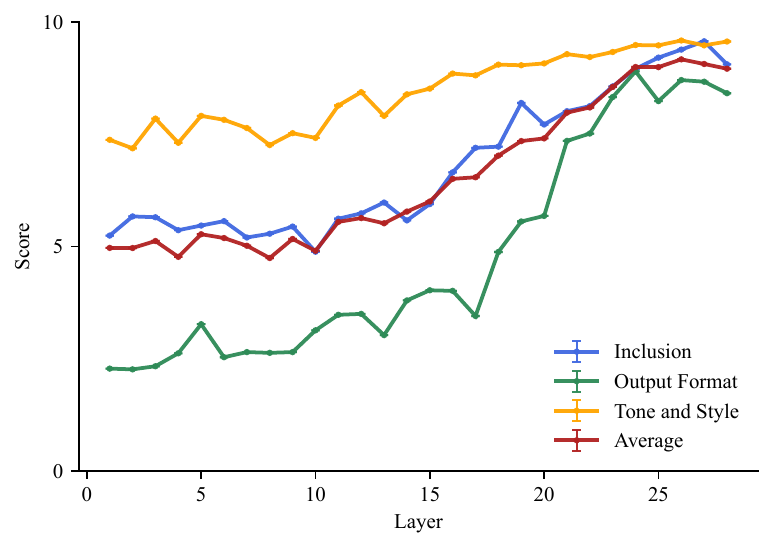}
  \caption{\label{trace}Layer-from context-masking analysis on \textsc{TRACE} benchmark. Each curve shows model performance (scored 0–10) on a distinct constraint type, evaluated by DeepSeek-V3.}
\end{figure}

\paragraph{Experiment.}
We select a subset of \textsc{TRACE} (69 prompts) that include all the following constraint types: \textbf{Inclusion}, \textbf{Output Format}, and \textbf{Tone and Style}. We use Qwen2.5-7B-Instruct \cite{yang2024qwen2} as the test model to complete the instructions, applying context-masking to the constraint tokens from each layer onward. To evaluate the quality of constraint satisfaction at each masking depth, we use DeepSeek-V3 \cite{liu2024deepseek} as an evaluator model, producing per-constraint scores from 0 to 10 (Prompt details can be found in Appendix \ref{appendix-prompt}).

Figure \ref{trace} shows the resulting line chart. We observe notable differences in the learning dynamics across constraint types:
\begin{itemize}
    \item \textbf{Output Format:} The learning curve is characterized by a sharp increase in score between layers 17–20, suggesting that formatting constraints (e.g., JSON structure, character encoding) are learned relatively late in the network.
    \item \textbf{Inclusion} and \textbf{Tone and Style:} These constraints show more gradual and smooth improvements across layers, indicating a slower or more distributed learning process.
\end{itemize}
These results demonstrate that different constraint types are learned at different depths in the model, further supporting our hypothesis of sequentially learning dynamics in composite instruction-following tasks.

\section{Related Work}
\paragraph{Multi-Hop Reasoning in LLMs.}
Recent studies have examined how large language models (LLMs) perform \textit{ latent factual multi-hop reasoning} \cite{press2023measuring, yang2024large2, li2024understanding, ju2024investigating}. \citet{yang2024large} finds that LLMs often reliably recall intermediate entities but inconsistently use them to complete complex prompts. \citet{biran2024hopping} shows that LLMs resolve intermediate entities early when answering multi-hop queries, and proposes a back-patching method to improve the performance. \citet{yu2025back} introduces logit flow to analyze latent multi-hop reasoning in LLMs and proposes back attention to improve accuracy. \citet{biology} identifies intermediate entities and reasoning path by Cross-Layer Transcoder. While those often focus on factual recall, our work investigates task-level reasoning rather than just chains of facts. A few recent studies have also investigated latent CoT reasoning \cite{wanglatent, chen2025reasoning, li2025implicit, mamidanna2025all}, in which inference is carried out within latent spaces.

\paragraph{Task Representations in ICL.}
The ability of LLMs to perform In-Context Learning (ICL) \cite{brown2020language} has spurred rich research into its internal mechanism. A prominent line of inquiry focuses on explicit task representations \cite{hendel2023context, liu2024context, toddfunction, liimplicit, saglam2025learning, yang2025task}. Initial work by \citet{hendel2023context} derived task vectors from layer activations. Other approaches include In-Context Vectors (ICVs) \cite{liu2024context} derived from principal components of activation differences, and Function Vectors (FVs) \cite{toddfunction} which emphasize the role of specific attention heads. While these foundational studies demonstrate how a singular task can be abstracted into a vector, our work extends this by investigating how composite tasks are handled. 

\paragraph{Mechanistic Interpretability.} Mechanistic interpretability \cite{elhage2021mathematical} aims to reverse engineer the internal mechanisms of LLMs. One type of studies focus on constructing the circuit in the model \cite{olsson2022context, wanginterpretability, gouldsuccessor, marks2024sparse}. Another line of work focuses on understanding intermediate representations through tools such as the LogitLens \cite{logitlen}. This technique has been extended to trace hidden states in LLMs \cite{dar2023analyzing,halawioverthinking, merullo2024language, wiegreffe2024answer}. Another major methodology is causal mediation analysis \cite{toddfunction, vig2020investigating, meng2022locating, geva2023dissecting, hendel2023context, wu2023depn, dumas2024separating}, which measures the effect of intervening on a hidden state to determine its causal contribution to the model's output. Recent work also investigates the superposition hypothesis \cite{elhage2022toy, scherlis2022polysemanticity}. To disentangle such representations, sparse autoencoders (SAEs) have been employed to extract interpretable features from high-dimensional activations \cite{gao2024scaling, marks2024sparse, saeclaude, ferrando2024know}.

\section{Conclusion}
We show that large language models (LLMs) exhibit an \textit{internal chain-of-thought}. Two claims ground our study: (i) distinct subtasks are learned at different network depths, and (ii) these subtasks are executed sequentially across layers. On a benchmark of 15 two-step composite tasks, we employ layer-from context-masking and propose a novel cross-task patching method, confirming (i). To examine claim (ii), we apply LogitLens to decode hidden states, revealing a consistent layerwise execution pattern. We further replicate our analysis on the real-world \textsc{Trace} benchmark, observing the same stepwise dynamics. Together, our results enhance LLMs transparency by showing their capacity to internally plan and execute subtasks (or instructions), opening avenues for fine-grained, instruction-level activation steering.

\section*{Acknowledgments}
This work was supported by the National Natural Science Foundation of China (Grant No.62506318); Guangdong Provincial Department of Education Project (Grant No.2024KQNCX028); CAAI-Ant Group Research Fund; Scientific Research Projects for the Higher-educational Institutions (Grant No.2024312096), Education Bureau of Guangzhou Municipality; Guangzhou-HKUST(GZ) Joint Funding Program (Grant No.2025A03J3957), Education Bureau of Guangzhou Municipality.

\section*{Limitations}
While our findings offer mechanistic insights into how LLMs internally decompose and execute composite tasks, several limitations must be acknowledged:

\paragraph{Task Construction Bias.}
The distinct ``X-shape'' pattern observed in our context-masking experiments (Figure \ref{mask_main}) is facilitated by the deliberate design of our benchmark tasks, which feature clearly distinguishable subtask types (e.g., knowledge retrieval followed by algorithmic transformation). This separation likely leads to more temporally distant ``learning points'' for each subtask across layers. However, when faced with a greater number or more nuanced types of subtasks, particularly those with high conceptual similarity, the context-masking technique might be less effective at clearly disentangling their individual learning stages. Indeed, as observed in our \textsc{Trace} analysis (Section \ref{section:trace}), the learning dynamics for closely related constraints can be more intertwined. Extending this analysis to more entangled, real-world tasks remains an important direction for future work.

\paragraph{\textsc{TRACE} Evaluation.}
Our \textsc{TRACE} evaluation relies on LLM-based scoring, which may suffer from calibration issues, prompt sensitivity, and lack of human ground truth; thus, we treat the results as exploratory. Alternatives such as logit-based measures, symbolic checks, or human validation could improve reliability, and we also acknowledge the lack of systematic error analysis. While symbolic evaluation can capture certain constraints, many \textsc{TRACE} tasks are inherently free-form, making LLM-based evaluation a flexible—if imperfect—choice.

\paragraph{Model and Scale Scope.} 
Our experiments are conducted on four mid-sized open-source models (3B–8B parameters). While these models are representative of common deployment settings, it remains an open question whether the observed phenomena generalize to larger frontier models (e.g., GPT-4, Claude). Differences in architecture, training corpus, and alignment objectives may yield distinct patterns of subtask representation or execution.

\bibliography{custom}

\clearpage

\appendix
\section{Datasets}
\label{appendix-dataset}
We evaluate our two central claims on 15 composite tasks spanning four categories. Each composite task is composed of two sequential subtasks, denoted as $s_1 \circ s_2$, and is designed to probe whether LLMs internally represent and apply these subtasks in a layered fashion. A summary of all composite tasks can be found in Table \ref{tab:dataset-summary}. We also describe the subtasks used in the cross-task patching experiment.

\paragraph{Antonym–Uppercase}
This composite dataset is constructed by capitalizing answers from an antonym dataset. The underlying antonym pairs are drawn from \citet{nguyen2017distinguishing}, which includes both antonyms and synonyms (e.g., \texttt{``good $\to$ bad''}). We follow the preprocessing procedure described in \citet{toddfunction}, then capitalize the antonym response, producing pairs like \texttt{``good $\to$ Bad''}. Intermediate answers are defined as the antonym in lowercase (e.g., \texttt{``bad''}) and the capitalized form of the query (e.g., \texttt{``Good''}).

\paragraph{Synonym–Uppercase}
Constructed in the same way as Antonym–Uppercase, using synonym pairs from \citet{nguyen2017distinguishing}. We capitalize the synonym to form the composite answer, and treat the lowercase synonym and capitalized query as intermediate outputs.

\paragraph{Country\_Capital–Lowercase}
This dataset is built from a country–capital mapping dataset \cite{toddfunction}. We lowercase the capital names to form the composite answers. For example, \texttt{``France $\to$ paris''}.

\paragraph{Landmark\_Country–Lowercase}
Pairs landmark names with their respective countries, based on data from \citet{hernandezlinearity}. The country name is lowercased to form the composite answer.

\paragraph{Product\_Company–Lowercase}
This dataset contains commercial products paired with the companies that produce them, also curated from \citet{hernandezlinearity}. The company name is lowercased to produce the final output.

\paragraph{Choose\_Last–Country\_Capital}
We use the country–capital dataset \cite{toddfunction} to create lists of three countries sampled at random. The final answer is the capital of the last country in the list. Intermediate outputs include the last country name and the capital of the first country.

\paragraph{Choose\_Last–Landmark\_Country}
Follows the same format as Choose\_Last–Country\_Capital, using landmark–country pairs from \citet{hernandezlinearity}. The model must extract the last landmark and map it to its corresponding country.

\paragraph{Adjective\_v\_Verb–Antonym}
This dataset tests syntactic category identification and semantic reasoning. From the antonym dataset \cite{nguyen2017distinguishing}, we select words that are unambiguously adjectives or verbs. Each list contains two verbs and one adjective. The model must identify the adjective and return its antonym.

\paragraph{Adjective\_v\_Verb–Synonym}
Constructed identically to Adjective\_v\_Verb–Antonym, but with synonym retrieval instead.

\paragraph{Choose\_Last–First\_Letter}
Constructed from a simple list-based selection dataset \cite{toddfunction}. The model is prompted with a list of three items and must return the first letter of the last item.

\paragraph{Choose\_Last–Uppercase}
Similar to Choose\_Last–First\_Letter, but instead of returning the first letter, the model is required to return the last item in uppercase form.

\paragraph{Antonym–English\_French}
We translate answers from the antonym dataset to French using the Google Translate API. The composite task consists of performing the antonym transformation and then translating the result. Intermediate answers include the English antonym and the French translation of the query.

\paragraph{Antonym–English\_Spanish}
Same as Antonym–English\_French, but translated to Spanish.

\paragraph{Landmark\_Country–English\_French}
Based on \citet{hernandezlinearity}, we first retrieve the country associated with a landmark, then translate the country name to French.

\paragraph{Landmark\_Country–English\_Spanish}
Constructed in the same way as Landmark\_Country–English\_French, using Spanish as the target language.

\begin{table*}[t]
  \centering
  \renewcommand{\arraystretch}{1.2}
  \setlength{\tabcolsep}{12pt}
  \begin{tabular}{ll}
    \hline
    \textbf{Category} & \textbf{Composite Tasks} \\
    \hline
    \textbf{Knowledge–Algorithmic} &
    \begin{tabular}[t]{@{}l@{}}
      Antonym–Uppercase \\
      Synonym–Uppercase \\
      Country\_Capital–Lowercase \\
      Landmark\_Country–Lowercase \\
      Product\_Company–Lowercase
    \end{tabular} \\
    \hline
    \textbf{Extractive–Knowledge} &
    \begin{tabular}[t]{@{}l@{}}
      Choose\_Last–Country\_Capital \\
      Choose\_Last–Landmark\_Country \\
      Adjective\_v\_Verb–Antonym \\
      Adjective\_v\_Verb–Synonym
    \end{tabular} \\
    \hline
    \textbf{Extractive–Algorithmic} &
    \begin{tabular}[t]{@{}l@{}}
      Choose\_Last–First\_Letter \\
      Choose\_Last–Uppercase
    \end{tabular} \\
    \hline
    \textbf{Knowledge–Translation} &
    \begin{tabular}[t]{@{}l@{}}
      Antonym–English\_French \\
      Antonym–English\_Spanish \\
      Landmark\_Country–English\_French \\
      Landmark\_Country–English\_Spanish
    \end{tabular} \\
    \hline
  \end{tabular}
  \caption{\label{tab:dataset-summary}
  Summary of the 15 composite tasks used in our experiments. Each task consists of a pair of subtasks ($s_1 \circ s_2$), spanning four categories: Knowledge–Algorithmic, Extractive–Knowledge, Extractive–Algorithmic, and Knowledge–Translation.
  }
\end{table*}

Below, we describe the individual subtasks ($s_1$ and $s_2$) used in the cross-task patching experiments. Each subtask is a functional unit that appears as part of one or more composite tasks.

\paragraph{Antonym} 
The antonym dataset is based on data from \citet{nguyen2017distinguishing}, which contains word pairs that are either antonyms or synonyms (e.g., \texttt{``good $\to$ bad''}, \texttt{``spirited $\to$ fiery''}). We follow the same preprocessing protocol as in \citet{toddfunction}.

\paragraph{Synonym} 
This dataset is also derived from \citet{nguyen2017distinguishing}, containing word pairs with synonym relationships. Preprocessing follows the same steps as the antonym dataset.

\paragraph{Country\_Capital} 
This dataset consists of country–capital pairs (e.g., \texttt{``France $\to$ Paris''}), taken from \citet{toddfunction}.

\paragraph{Landmark\_Country} 
Includes landmark–country pairs such as \texttt{``Eiffel Tower $\to$ France''}, based on the dataset from \citet{hernandezlinearity}.

\paragraph{Product\_Company} 
Contains entries mapping commercial products to the companies that produce or sell them (e.g., \texttt{``iPhone $\to$ Apple''}). Also sourced from \citet{hernandezlinearity}.

\paragraph{Choose\_Last} 
Constructed by sampling three items and asking the model to return the last item. Data sourced from \citet{toddfunction}.

\paragraph{Adjective\_v\_Verb} 
This dataset is designed to test part-of-speech reasoning. Each example contains a list of two verbs and one adjective, and the model must identify the adjective. Source: \citet{toddfunction}.

\paragraph{Uppercase} 
A simple string transformation task where the model is required to convert the input to uppercase. Examples and format are adapted from \citet{toddfunction}.

\paragraph{Lowercase} 
Analogous to the Uppercase task, but the model is required to convert the input to lowercase. Based on the same dataset used in \citet{toddfunction}.

\paragraph{First\_Letter} 
The task involves selecting the first letter of a given word. We construct this by reusing inputs from the Uppercase dataset and extracting only the first character.

\paragraph{Translation (English–French / English–Spanish)} 
We use bilingual word pairs from \citet{conneau2017word} for English–French and English–Spanish translations. Each example consists of an English word and its corresponding translation. We follow the preprocessing pipeline used in \citet{toddfunction}.

\section{Prompt Details}
\label{appendix-prompt}
\textit{In-context learning prompt:}
\begin{tcolorbox}[breakable]
Q: \{$x_{1}$\}\\
A: \{$y_{1}$\}\\
\\
Q: \{$x_{2}$\}\\
A: \{$y_{2}$\}\\ \\Q: \{$x_{3}$\}\\A: \{$y_{3}$\}\\ \\Q: \{$x_{4}$\}\\A: \{$y_{4}$\}\\ \\Q: \{$x_{5}$\}\\A: \{$y_{5}$\}\\ \\Q: \{$x_{q}$\}\\A:
\end{tcolorbox}

\textit{An example (Chinese version) of \textsc{TRACE} dataset:}
\begin{CJK}{UTF8}{gbsn}
\begin{tcolorbox}[breakable]
\textbf{Task Description:} 详细解释Adagrad算法，扮演一位机器学习专家的角色。\\
\textbf{Constraints:} 
1. 至少写1000字； \\
2. 解释中要包含Adagrad算法的起源、原理、优缺点以及在实际场景中的应用；\\
3. 使用LaTeX格式来表示所有数学公式和算法步骤；\\
4. 确保原理解释既专业又易于理解。
\end{tcolorbox}
\end{CJK}

\textit{LLM evaluation prompt (TRACE):}
\begin{tcolorbox}[breakable]
\text{[System]}\\
You are a fair judge, and please evaluate the quality of an AI assistant’s responses to user query. You need to assess the response based on the following constraints. We will provide you with the user’s query, some constraints, and the AI assistant’s response that needs your evaluation. When you commence your evaluation, you should follow the following process:\\
1. Evaluate each constraint: Assess how well the AI assistant’s response meets each individual constraint.\\
2. Assign a score (0–10) for each constraint: After explaining your assessment for each constraint, give a corresponding score from 0 (does not meet the requirement at all) to 10 (fully meets the requirement).\\
3. List the scores: List the Constraints Overall Score (as a list of the individual scores in their original constraint order).\\
4. Strict scoring policy: Be as strict as possible in assigning scores. If the response is irrelevant, contains major factual errors, or generates harmful content, the “Constraints Overall Score” must be 0.\\
5. Preserve constraint order: When you provide the “Fine Grained Score,” the constraints must appear in the same order as they are listed in the input context.\\
6. Follow the output format: After you provide explanations for each constraint, list the Fine Grained Score in JSON format and the Constraints Overall Score as a list, as shown in the example below.\\
Please reference and follow the format demonstrated in the /* Example */.\\
/* Example */\\
—INPUT—\\
\text{\#Task Description:}\\
Create a password for this account\\
\text{\#Constraints:}\\
The password must be at least 8 characters long;\\
It must contain 1 uppercase letter;\\
It must contain 1 lowercase letter;\\
It must include 2 numbers;\\
\#Input:\\
NULL\\
\#Response:\\
Ax7y4gTf\\
—OUTPUT—\\
Explanation:\\
Password Length: The password “Ax7y4gTf” is 8 characters long, meeting the first constraint, scoring 10 points.\\
Contains 1 uppercase letter: The password “Ax7y4gTf” contains two uppercase letters, “A” and “T”, which means it meets the second constraint, but the explanation incorrectly states it does not meet the constraint, scoring 0 points.\\
Contains 1 lowercase letter: The password “Ax7y4gTf” contains three lowercase letters, “x”, “y”, and “g”, which means it meets the third constraint, but the explanation incorrectly states it does not meet the constraint, scoring 0 points.\\
Includes 2 numbers: The password “Ax7y4gTf” includes two numbers, “7” and “4”, meeting the fourth constraint, scoring 10 points.\\
Fine Grained Score: [ \{ "The password must be at least 8 characters long": 10, "It must contain 1 uppercase letter": 0, "It must contain 1 lowercase letter": 0, "It must include 2 numbers": 10 \} ]\\
Constraints Overall Score: [10, 0, 0, 10]\\
/* Input */\\
—INPUT—\\
\#Task Description:\\
\{task\_description\}\\
\#Constraints:\\
\{constraint\}\\
\#Input:\\
\{input\}\\
<response>:\\
\{ans\}\\
—OUTPUT—
\end{tcolorbox}
We extract the ``Constraints Overall Score'' from the last line of the evaluator's response using template matching. To compute the final score per constraint type, we average over the number of relevant constraints (not samples).

\section{Causal Intervention Experiment for Claim 2}
\label{appendix-causal}
While LogitLens provides valuable correlational evidence, it does not establish causal dependence between subtask computations. To address this limitation, we performed an additional causal intervention experiment, targeting the residual stream at the peak layer of subtask $s_1$, as identified in our decoding analysis (e.g., Figure \ref{decoding_res_L1}).

Given a prompt $p$ with intermediate and final answers $y^{s_1}$, $y^{s_2}$, $y^t$, and an alternative prompt $x^*$ with corresponding answers $y^{*s_1}$, $y^{*s_2}$, $y^{*t}$, we performed the following intervention at layer $l^*$ (peak layer for subtask $s_1$):
\begin{equation}
    \mathbf{h}^{l^*} \leftarrow \mathbf{h}^{l^*} + \mathbf{h}^{l^*} {\mathbf{W_U}^{y^{s_1}}}^\top (\mathbf{W_U}^{y^{*s_1}} - \mathbf{W_U}^{y^{s_1}})
\end{equation}
This operation swaps out the residual representation aligned with $y^{s_1}$ and replaces it with that of $y^{*s_1}$, without changing other components of the prompt.

We then measured the Mean Reciprocal Rank (MRR) of the final outputs for both $y^t$ and $y^{*t}$. The results, evaluated on all 15 tasks using Llama-3.1-8B, show that in most cases, this substitution causes the model's prediction to shift from $y^t$ to $y^{*t}$ (see Table \ref{table-causal}). This provides causal evidence that the output of the first subtask directly influences the computation of the second, supporting our claim that subtasks are executed sequentially in the form of $s_2(s_1(x))$.

\begin{table*}
\footnotesize 
  \centering
  \begin{tabular}{lllll}
    \hline

    \textbf{Composite Task} & $y^t$ \textbf{(base)} & $y^{*t}$ \textbf{(base)} & $y^t$ \textbf{(intervention)} & $y^{*t}$ \textbf{(intervention)}  \\
    \hline
    \texttt{antonym-uppercase} 	&0.76&0.01&0.05 (\textbf{-0.71})&0.54 (\textbf{+0.53})\\

    \texttt{synonym-uppercase} &0.58&0.01&0.04 (\textbf{-0.54})&0.44 (\textbf{+0.43}) \\

    \texttt{country\_capital-lowercase} &0.95&0.02&0.06 (\textbf{-0.89})&0.34 (\textbf{+0.32})  \\

    \texttt{landmark\_country-lowercase} &0.93&0.07&0.05 (\textbf{-0.88})&0.87 (\textbf{+0.80})  \\

    \texttt{product\_company-lowercase} &0.84&0.16&0.12 (\textbf{-0.72})&0.70 (\textbf{+0.54})  \\

    \texttt{choose\_last-country\_capital} &0.96&0.03&0.42 (\textbf{-0.54})&0.21 (\textbf{+0.18})   \\

    \texttt{choose\_last-landmark\_country} &0.86&0.12&0.59 (\textbf{-0.27})&0.10 (\textbf{+0.02})  \\

    \texttt{adjective\_v\_verb-antonym} &0.43&0.01&0.14 (\textbf{-0.29})&0.03 (\textbf{+0.02})   \\

    \texttt{adjective\_v\_verb-synonym} &0.41&0.01&0.05 (\textbf{-0.36})&0.09 (\textbf{+0.08})   \\

    \texttt{choose\_last-first\_letter} &0.88&0.15&0.74 (\textbf{-0.14})&0.37 (\textbf{+0.22})  \\

    \texttt{choose\_last-uppercase} &0.96&0.01&0.37 (\textbf{-0.59})&0.41 (\textbf{+0.40})   \\

    \texttt{antonym-english\_french} &0.53&0.01&0.04 (\textbf{-0.49})&0.35 (\textbf{+0.34})   \\

    \texttt{antonym-english\_spanish} &0.53&0.01&0.05 (\textbf{-0.48})&0.35 (\textbf{+0.34})   \\

    \texttt{landmark\_country-english\_french} &0.91&0.08&0.09 (\textbf{-0.82})&0.66 (\textbf{+0.58})  \\

    \texttt{landmark\_country-english\_spanish} &0.88&0.06&0.05 (\textbf{-0.83})&0.63 (\textbf{+0.57})  \\

    \textbf{Average} &0.76&0.05&0.19 (\textbf{-0.57})&0.41 (\textbf{+0.36})   \\

    \hline
  \end{tabular}
  \caption{\label{table-causal}Causal intervention results for all composite tasks in Llama-3.1-8B.}
\end{table*}

\section{Results of Layer-from Context-Masking}
\label{appendix-masking}

We present the complete results of the Layer-from Context-Masking experiments across four models. Each figure visualizes the layer-wise performance on all 15 composite tasks, showing how masking context information from progressively later layers affects the model's ability to complete subtasks and composite outputs.

\begin{figure*}[htbp]
	\centering
	\includegraphics[width=16cm]{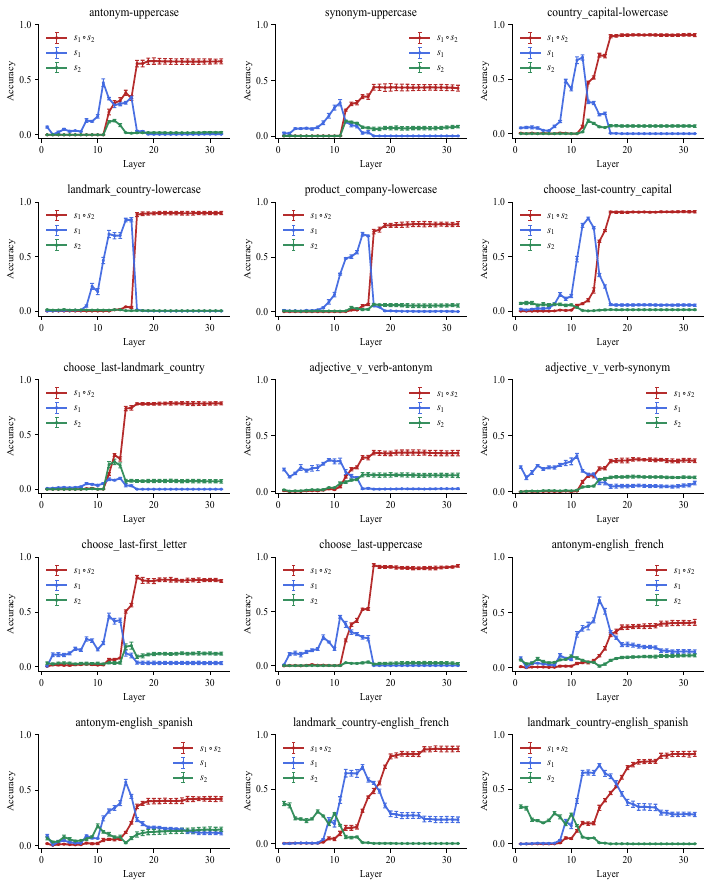}
	\caption{\label{masking_L1}Layer-from context-masking results for all composite tasks in Llama-3.1-8B.}
\end{figure*}

\begin{figure*}[htbp]
	\centering
	\includegraphics[width=16cm]{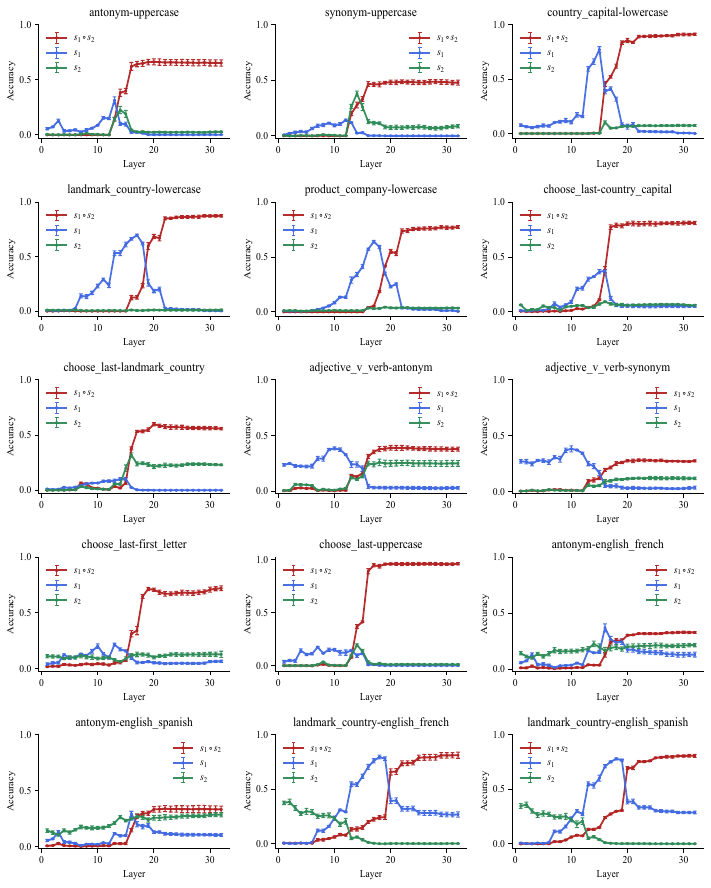}
	\caption{\label{masking_M}Layer-from context-masking results for all composite tasks in Mistral-7B.}
\end{figure*}

\begin{figure*}[htbp]
	\centering
	\includegraphics[width=16cm]{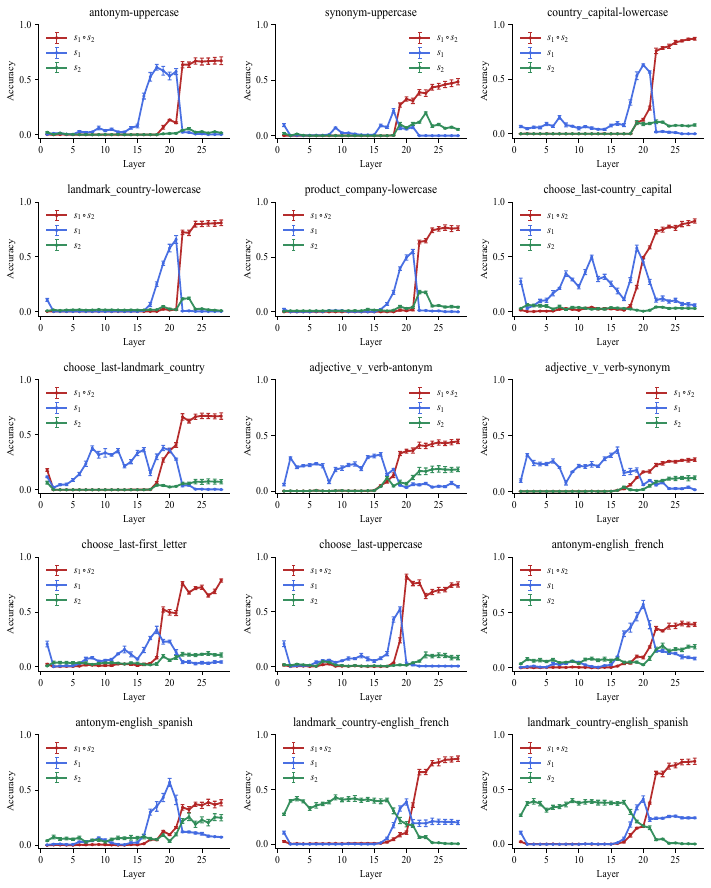}
	\caption{\label{masking_Q}Layer-from context-masking results for all composite tasks in Qwen2.5-7B.}
\end{figure*}

\begin{figure*}[htbp]
	\centering
	\includegraphics[width=16cm]{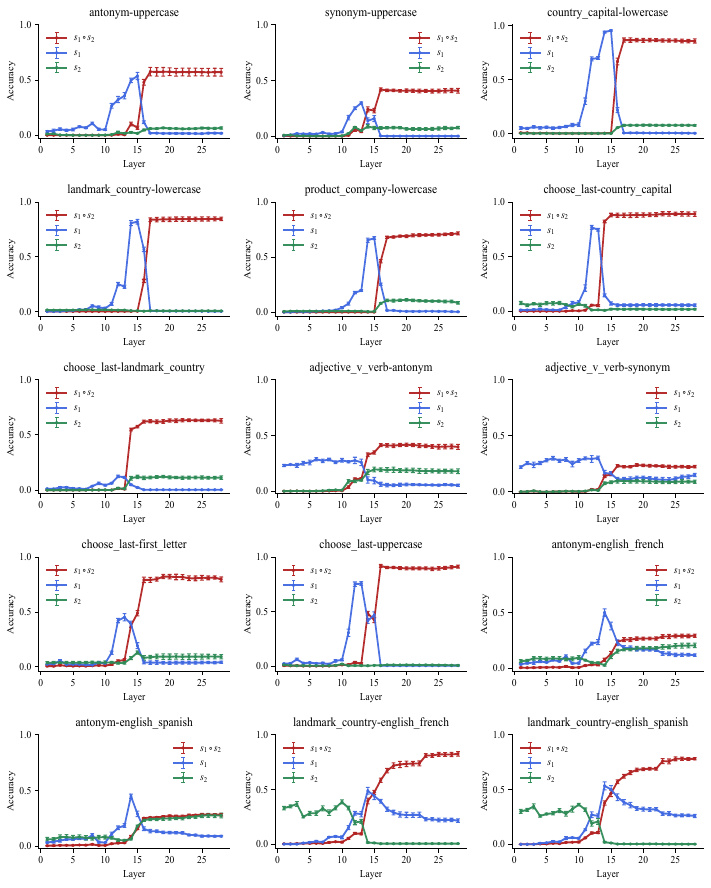}
	\caption{\label{masking_L2}Layer-from context-masking results for all composite tasks in Llama-3.2-3B.}
\end{figure*}

\begin{itemize}
    \item Figure \ref{masking_L1} shows results for Llama-3.1-8B.
    \item Figure \ref{masking_M} shows results for Mistral-7B.
    \item Figure \ref{masking_Q} shows results for Qwen2.5-7B.
    \item Figure \ref{masking_L2} shows results for Llama-3.2-3B.
\end{itemize}

\section{Results of Cross-Task Patching}
\label{appendix-patching}

We report the full results of the Cross-Task Patching experiment across all four models. Table \ref{strength_all} summarizes the subtask vector strength for each model, indicating how well activations from composite tasks can transfer to individual subtasks.

\section{Results of Logit Decoding}
\label{appendix-decoding}

We present the complete results of the Logit Decoding analysis for all four models. Each model has two figures:
(1) Mean Reciprocal Rank (MRR) scores of component outputs (attention and MLP layers), and 
(2) Mean Reciprocal Rank (MRR) scores of residual stream.

\begin{figure*}[htbp]
	\centering
	\includegraphics[width=16cm]{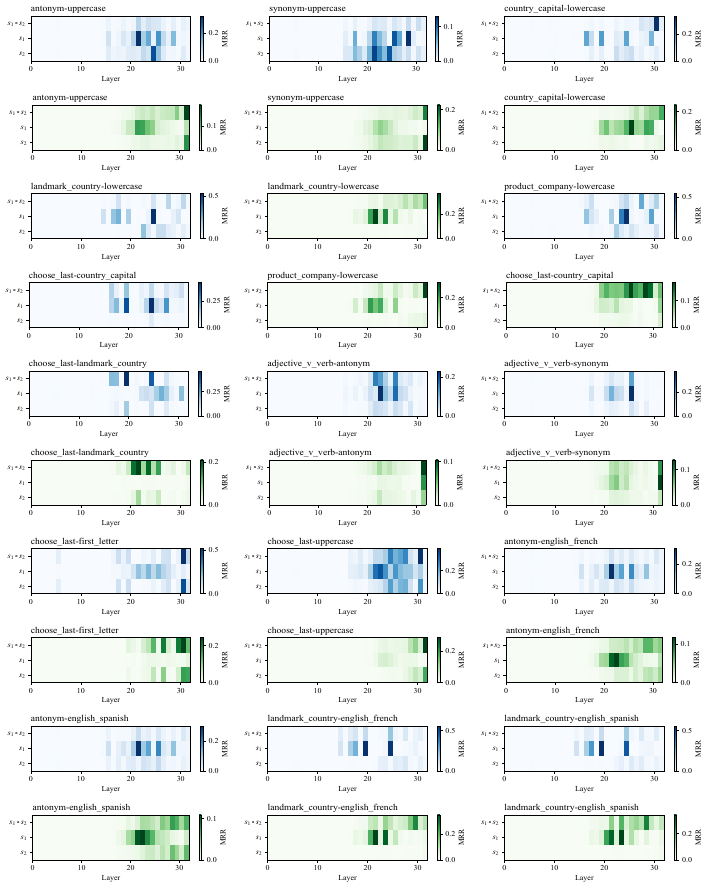}
	\caption{\label{decoding_com_L1}Heatmaps of attention and MLP block decoding results for all tasks in Llama-3.1-8B.}
\end{figure*}

\begin{figure*}[htbp]
	\centering
	\includegraphics[width=16cm]{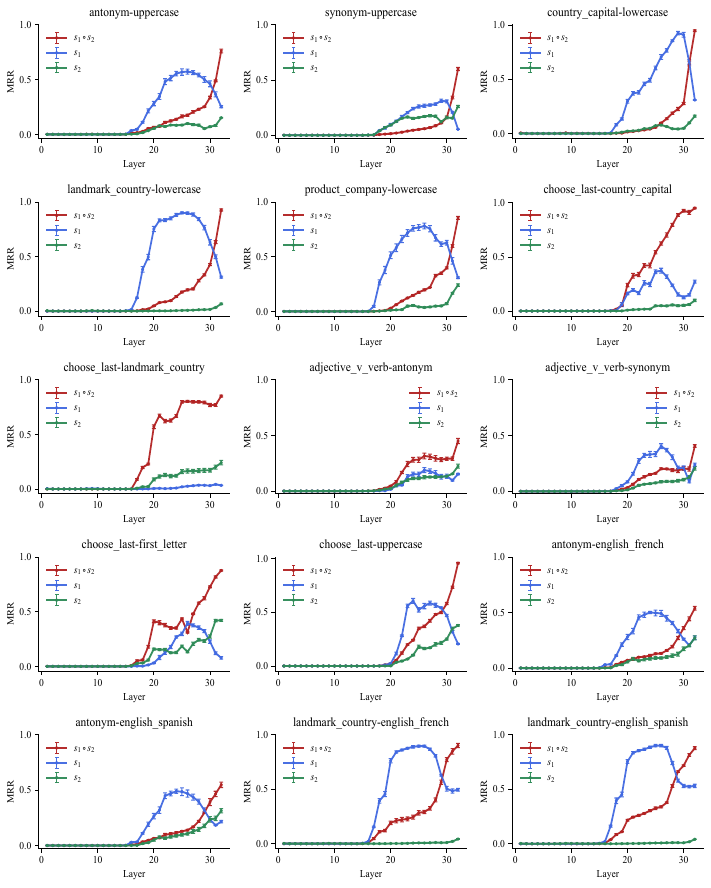}
	\caption{\label{decoding_res_L1}Decoding results of residual stream for all tasks in Llama-3.1-8B.}
\end{figure*}

\begin{figure*}[htbp]
	\centering
	\includegraphics[width=16cm]{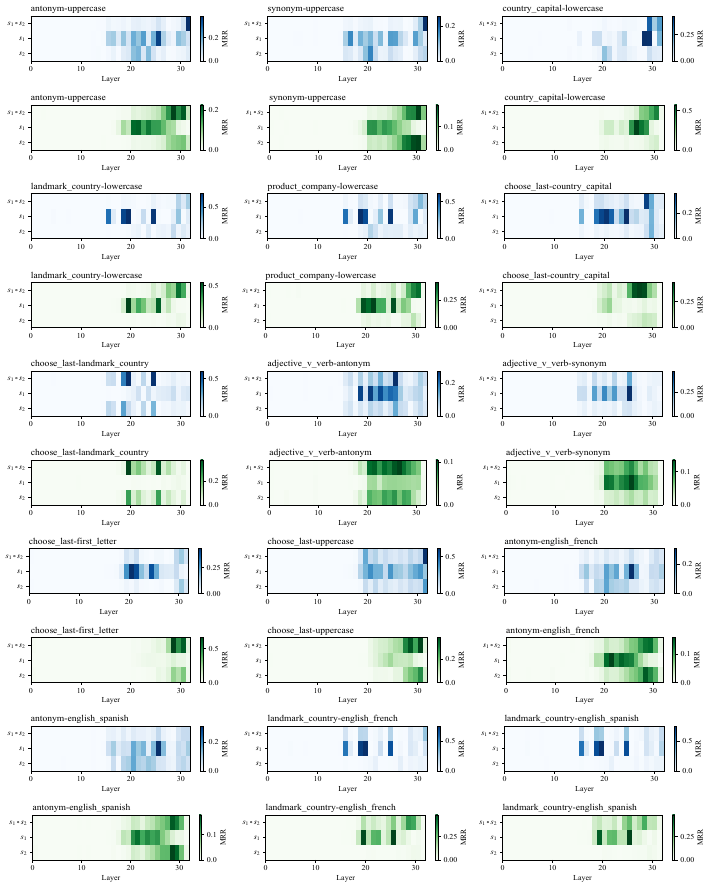}
	\caption{\label{decoding_com_M}Heatmaps of attention and MLP block decoding results for all tasks in Mistral-7B.}
\end{figure*}

\begin{figure*}[htbp]
	\centering
	\includegraphics[width=16cm]{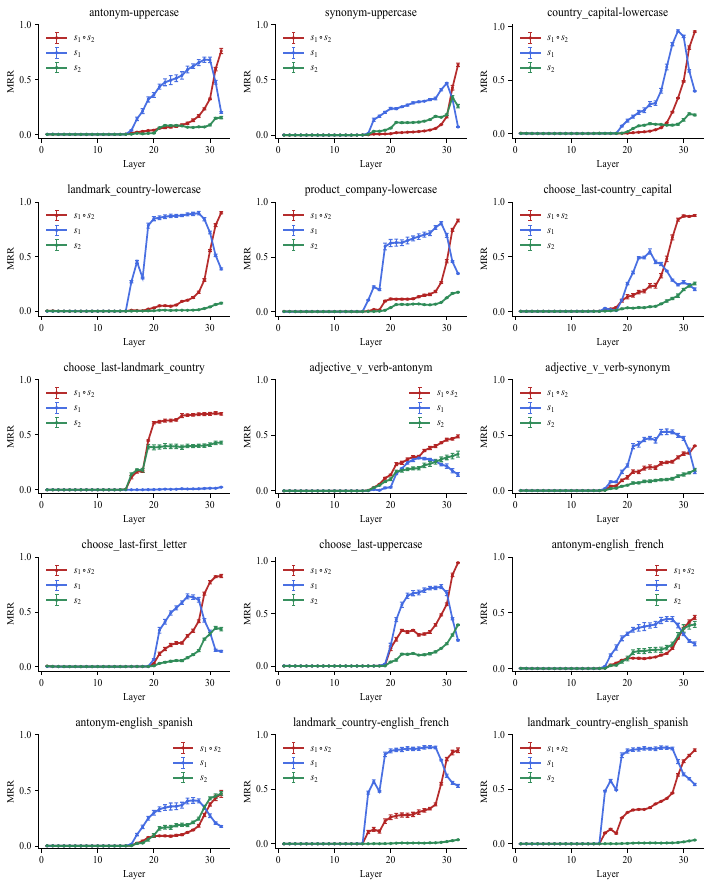}
	\caption{\label{decoding_res_M}Decoding results of residual stream for all tasks in Mistral-7B.}
\end{figure*}

\begin{figure*}[htbp]
	\centering
	\includegraphics[width=16cm]{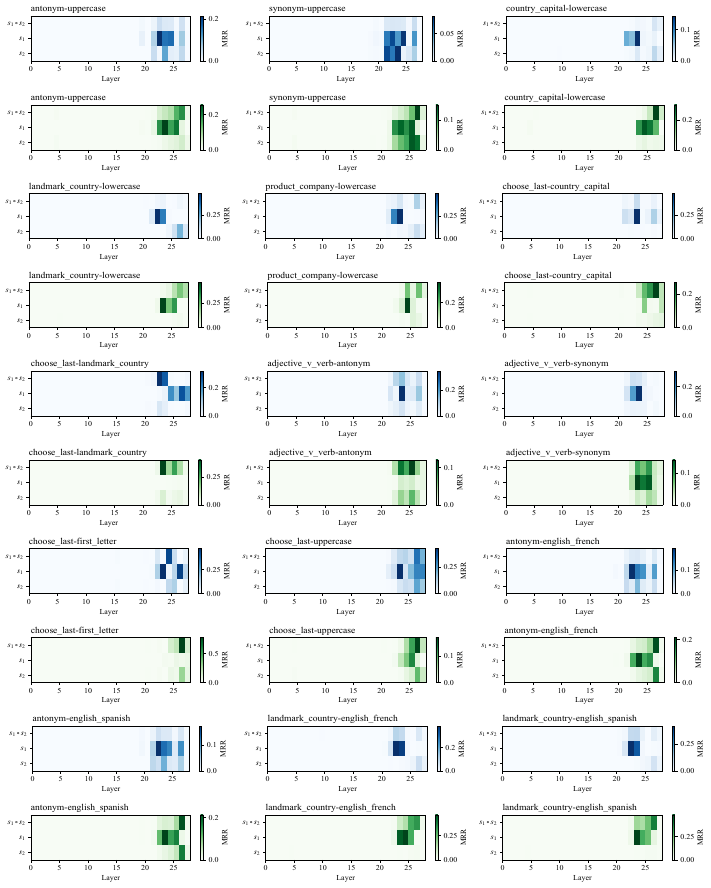}
	\caption{\label{decoding_com_Q}Heatmaps of attention and MLP block decoding results for all tasks in Qwen2.5-7B.}
\end{figure*}

\begin{figure*}[htbp]
	\centering
	\includegraphics[width=16cm]{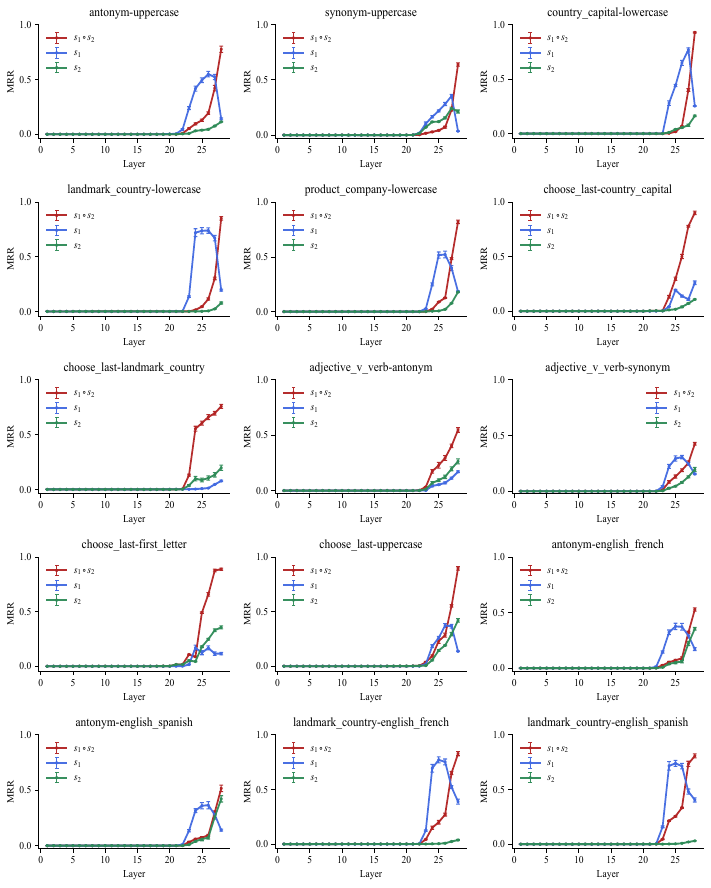}
	\caption{\label{decoding_res_Q}Decoding results of residual stream for all tasks in Qwen2.5-7B.}
\end{figure*}

\begin{figure*}[htbp]
	\centering
	\includegraphics[width=16cm]{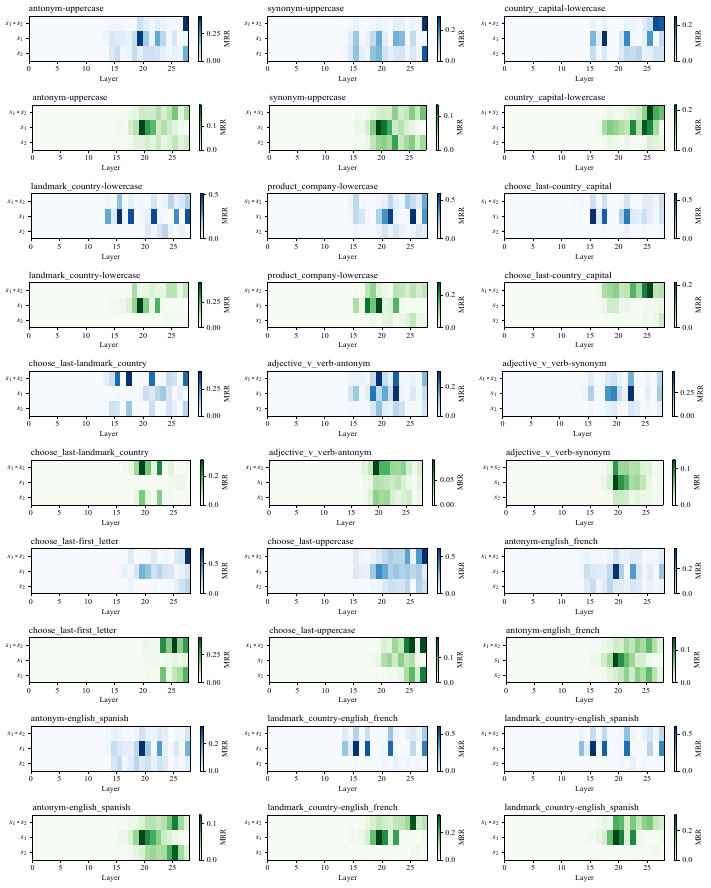}
	\caption{\label{decoding_com_L2}Heatmaps of attention and MLP block decoding results for all tasks in Llama-3.2-3B.}
\end{figure*}

\begin{figure*}[htbp]
	\centering
	\includegraphics[width=16cm]{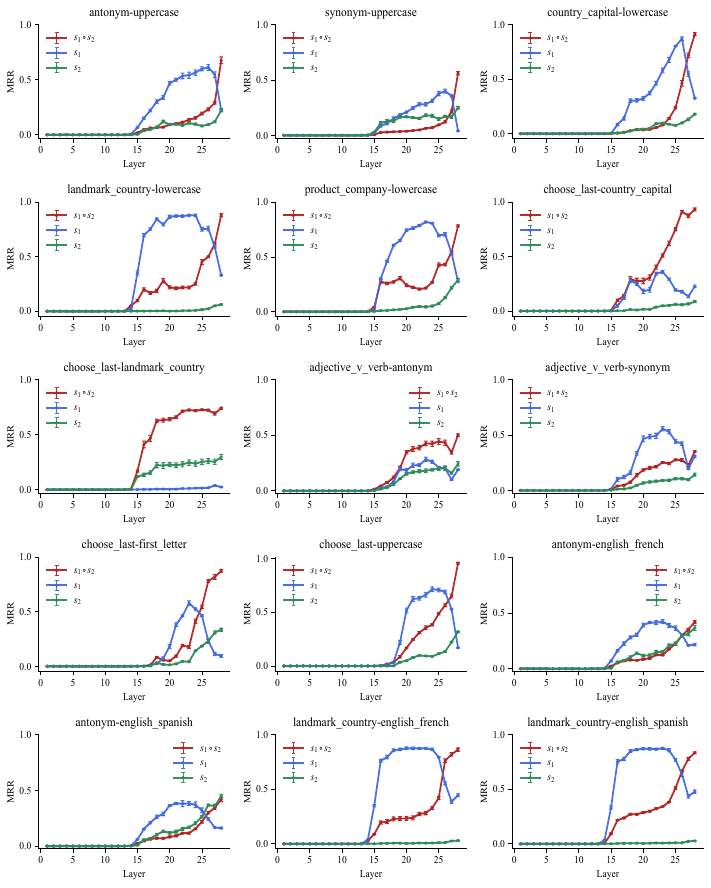}
	\caption{\label{decoding_res_L2}Decoding results of residual stream for all tasks in Llama-3.2-3B.}
\end{figure*}

\begin{itemize}
    \item Figures \ref{decoding_com_L1} and \ref{decoding_res_L1} show results for Llama-3.1-8B.
    \item Figures \ref{decoding_com_M} and \ref{decoding_res_M} show results for Mistral-7B.
    \item Figures \ref{decoding_com_Q} and \ref{decoding_res_Q} show results for Qwen2.5-7B.
    \item Figures \ref{decoding_com_L2} and \ref{decoding_res_L2} show results for Llama-3.2-3B.
\end{itemize}


\begin{table*}
\footnotesize                             % ↓ font size
\setlength{\tabcolsep}{6pt}               % ↓ horizontal padding (default ≈ 6pt)
  \centering
  \begin{tabular}{lllll}
    \hline
    \textbf{Composite Task} & \textbf{Llama-3.1-8B} & \textbf{Mistral-7B} & \textbf{Qwen2.5-7B} & \textbf{Llama-3.2-3B}  \\
    \hline
    \texttt{antonym-uppercase} &  & & & \\
    \quad $s_1$ (antonym) & 0.92 $\pm$ 0.02 & 0.26 $\pm$ 0.07 & 0.83 $\pm$ 0.02 & 0.87 $\pm$ 0.05 \\
    \quad $s_2$ (uppercase) & 0.24 $\pm$ 0.03 &0.35 $\pm$ 0.02 &0.03 $\pm$ 0.02 &0.34 $\pm$ 0.04\\

    \texttt{synonym-uppercase} &  & & & \\
    \quad $s_1$ (synonym) & 0.90 $\pm$ 0.05 &0.06 $\pm$ 0.03 &0.39 $\pm$ 0.02 &0.82 $\pm$ 0.04 \\
    \quad $s_2$ (uppercase) & 0.19 $\pm$ 0.05 &0.40 $\pm$ 0.03 &0.09 $\pm$ 0.03 &0.11 $\pm$ 0.02\\

    \texttt{country\_capital-lowercase} &  & & & \\
    \quad $s_1$ (country\_capital) & 0.88 $\pm$ 0.03 &0.20 $\pm$ 0.03 &0.49 $\pm$ 0.05 &1.02 $\pm$ 0.02 \\
    \quad $s_2$ (lowercase) & 0.49 $\pm$ 0.05 &0.45 $\pm$ 0.08 &0.34 $\pm$ 0.03 &0.31 $\pm$ 0.03\\

    \texttt{landmark\_country-lowercase} &  & & & \\
    \quad $s_1$ (landmark\_country) & 0.96 $\pm$ 0.02 &0.92 $\pm$ 0.02 &0.67 $\pm$ 0.02 &0.96 $\pm$ 0.01 \\
    \quad $s_2$ (lowercase) & 0.07 $\pm$ 0.03 &0.11 $\pm$ 0.04 &0.17 $\pm$ 0.04 &0.01 $\pm$ 0.00\\

    \texttt{product\_company-lowercase} &  & & & \\
    \quad $s_1$ (product\_company) & 1.01 $\pm$ 0.01 &0.86 $\pm$ 0.03 &0.65 $\pm$ 0.02 &0.99 $\pm$ 0.03 \\
    \quad $s_2$ (lowercase) & 0.05 $\pm$ 0.01 &0.46 $\pm$ 0.04 &0.36 $\pm$ 0.05 &0.06 $\pm$ 0.03\\

    \texttt{choose\_last-country\_capital} &  & & & \\
    \quad $s_1$ (choose\_last) & 0.44 $\pm$ 0.03 &0.32 $\pm$ 0.03 &0.37 $\pm$ 0.05 &0.26 $\pm$ 0.04 \\
    \quad $s_2$ (country\_capital) & 0.98 $\pm$ 0.02 &0.99 $\pm$ 0.01 &0.97 $\pm$ 0.01 &0.99 $\pm$ 0.01\\

    \texttt{choose\_last-landmark\_country} &  & & & \\
    \quad $s_1$ (choose\_last) & 0.03 $\pm$ 0.02 &0.00 $\pm$ 0.00 &0.05 $\pm$ 0.04 &0.03 $\pm$ 0.01 \\
    \quad $s_2$ (landmark\_country) & 0.94 $\pm$ 0.01 &0.96 $\pm$ 0.01 &0.91 $\pm$ 0.01 &0.96 $\pm$ 0.01\\

    \texttt{adjective\_v\_verb-antonym} &  & & & \\
    \quad $s_1$ (adjective\_v\_verb) & 0.38 $\pm$ 0.11 &0.21 $\pm$ 0.05 &0.54 $\pm$ 0.08 &0.16 $\pm$ 0.04 \\
    \quad $s_2$ (antonym) & 0.94 $\pm$ 0.03 &0.96 $\pm$ 0.04 &0.95 $\pm$ 0.01 &0.91 $\pm$ 0.05\\

    \texttt{adjective\_v\_verb-synonym} &  & & & \\
    \quad $s_1$ (adjective\_v\_verb) & 0.29 $\pm$ 0.04 &0.39 $\pm$ 0.05 &0.26 $\pm$ 0.07 &0.17 $\pm$ 0.04 \\
    \quad $s_2$ (synonym) & 0.99 $\pm$ 0.05 &0.76 $\pm$ 0.05 &0.83 $\pm$ 0.07 &0.79 $\pm$ 0.02\\

    \texttt{choose\_last-first\_letter} &  & & & \\
    \quad $s_1$ (choose\_last) & 0.27 $\pm$ 0.03 &0.08 $\pm$ 0.02 &0.25 $\pm$ 0.04 &0.41 $\pm$ 0.02 \\
    \quad $s_2$ (first\_letter) & 1.01 $\pm$ 0.01 &0.54 $\pm$ 0.03 &0.95 $\pm$ 0.05 &0.97 $\pm$ 0.03\\

    \texttt{choose\_last-uppercase} &  & & & \\
    \quad $s_1$ (choose\_last) & 0.44 $\pm$ 0.03 &0.12 $\pm$ 0.02 &0.26 $\pm$ 0.04 &0.56 $\pm$ 0.03 \\
    \quad $s_2$ (uppercase) & 0.99 $\pm$ 0.00 &0.99 $\pm$ 0.00 &1.00 $\pm$ 0.00 &0.98 $\pm$ 0.01\\

    \texttt{antonym-english\_french} &  & & & \\
    \quad $s_1$ (antonym) & 0.92 $\pm$ 0.02 &0.53 $\pm$ 0.05 &0.88 $\pm$ 0.02 &0.76 $\pm$ 0.07 \\
    \quad $s_2$ (english\_french) & 0.39 $\pm$ 0.04 &0.58 $\pm$ 0.05 &0.45 $\pm$ 0.01 &0.60 $\pm$ 0.06\\

    \texttt{antonym-english\_spanish} &  & & & \\
    \quad $s_1$ (antonym) & 0.91 $\pm$ 0.02 &0.35 $\pm$ 0.05 &0.86 $\pm$ 0.02 &0.70 $\pm$ 0.08 \\
    \quad $s_2$ (english\_spanish) & 0.45 $\pm$ 0.03 &0.69 $\pm$ 0.02 &0.57 $\pm$ 0.03 &0.70 $\pm$ 0.01\\

    \texttt{landmark\_country-english\_french} &  & & & \\
    \quad $s_1$ (landmark\_country) & 0.93 $\pm$ 0.01 &0.96 $\pm$ 0.01 &0.92 $\pm$ 0.01 &0.87 $\pm$ 0.01 \\
    \quad $s_2$ (english\_french) & 0.02 $\pm$ 0.00 &0.02 $\pm$ 0.01 &0.02 $\pm$ 0.01 &0.00 $\pm$ 0.00\\

    \texttt{landmark\_country-english\_spanish} &  & & & \\
    \quad $s_1$ (landmark\_country) & 0.93 $\pm$ 0.01 &0.95 $\pm$ 0.01 &0.92 $\pm$ 0.01 &0.90 $\pm$ 0.01 \\
    \quad $s_2$ (english\_spanish) & 0.01 $\pm$ 0.00 &0.01 $\pm$ 0.01 &0.01 $\pm$ 0.01 &0.00 $\pm$ 0.00\\
    \hline
  \end{tabular}
  \caption{\label{strength_all}Subtask vector strength for all composite tasks across four models.}
\end{table*}


\end{document}